\documentclass[11pt]{article}

\usepackage[final]{acl}

\usepackage{times}
\usepackage{latexsym}

\usepackage[T1]{fontenc}

\usepackage[utf8]{inputenc}

\usepackage{microtype}

\usepackage{inconsolata}

\usepackage{graphicx}
\usepackage{enumitem}
\usepackage{booktabs}   
\usepackage{url}        
\usepackage{hyperref}   
\usepackage{tabularx}
\usepackage{ragged2e} 
\usepackage{arydshln}
\usepackage{caption}
\usepackage{amsmath}

\usepackage{xcolor}
\definecolor{linkpink}{RGB}{255,0,170}
\author{
 \textbf{Mengzhang Li\textsuperscript{1}}\quad
 \textbf{Yuan Yao\textsuperscript{1,2}}
\\
 \textsuperscript{1}Shanghai Qizhi Institute,
 \textsuperscript{2}College of AI, Tsinghua University
\\
 \small{
   \textbf{Correspondence:} \href{mailto:yaoyuanthu@gmail.com}{yaoyuanthu@gmail.com}
 }
\\[0.5ex]
   \href{https://github.com/OpenSQZ/OpenGlass}{\textcolor{linkpink}{\texttt{https://github.com/OpenSQZ/OpenGlass}}
}
}
\title{OpenGlass: A Sensing-Computing Split Architecture for Local MLLM-Driven Real-Time Visual Assistance}

\begin{document}
\maketitle
\begin{abstract}
We present OpenGlass, an open-source, privacy-oriented, local-first system for low-latency multimodal visual assistance, with a primary focus on blind and low-vision users. Cloud MLLM assistants offer strong visual understanding, but often require uploading first-person visual data and can suffer multi-second network delays; wearable glasses are ideal for sensing, but cannot host large models under tight compute and power budgets. OpenGlass addresses this gap with a sensing-computing split: an ESP32-based glasses-side unit captures visual context, while a nearby consumer-grade device performs local MLLM inference and local speech output, reducing cloud reliance and keeping raw egocentric visual data on user-controlled devices by default. We evaluate response quality, query-ready-to-audio latency, safety-aware abstention, and auditable logs. Under real ESP32 Wi-Fi capture, OpenGlass reaches 993 ms median user-to-audio latency with resized payloads and 1625 ms with raw 1280$\times$720 payloads; 97.5\% and 93.3\% of trials fall below 2 s, respectively. OpenGlass is a user-initiated visual-assistance reference platform for obstacle/hazard awareness, sign/object queries, and image-quality self-checking, rather than a certified navigation aid. We release source code, hardware instructions, prompts, evaluation data, and logs.
\end{abstract}

\section{Introduction}
Multimodal foundation models have demonstrated strong visual understanding, yet deploying such capability in assistive settings still faces a fundamental tension between capability and interaction quality.
Assistive applications for blind and low-vision (BLV) users exemplify this gap: a wearable glasses system must capture the scene, interpret the user request, and deliver spoken guidance quickly enough for the user to act; otherwise, responses become stale in dynamic tasks such as item recognition and text reading ~\cite{bigham2010vizwiz}.
Today, many widely used visual assistance workflows and recent GPT class assistants are deployed as cloud services ~\cite{huang2025my}, which can provide strong model capability but often incur multi-second delays and network jitter under real wireless conditions. Cloud-only deployment also raises privacy concerns for egocentric assistive cameras, since first-person images may contain bystanders, private spaces, screens, documents, and other sensitive visual context.
At the same time, wearable devices are ideal for continuous sensing, yet they typically have limited compute and power budgets and therefore cannot host large multimodal models.

From a deployment perspective, existing compute platforms can be grouped into several categories.
Cloud clusters offer the highest compute capacity but depend on network availability and introduce variable transfer delays.
Local desktop workstations can run large models with stable throughput, but they are not naturally portable for everyday assistive use.
Portable consumer devices such as laptops, tablets, and phones provide a practical middle ground for running models locally near the user.
Wearable devices such as watches and glasses are effective for sensing, but are constrained in memory, compute, and battery.
In addition, many users understandably prefer not to continuously upload first-person camera images or spoken interaction content to external servers, since these streams may include private spaces, bystanders, screens, documents, and other sensitive context. A local-first design therefore serves not only as a latency optimization, but also as a privacy-oriented deployment choice: raw visual inputs and generated spoken feedback can remain on user-controlled devices by default, while cloud services are treated as optional baselines or online enhancements rather than the default execution path.

We present \textsc{OpenGlass}, an open-source, privacy-oriented system for low-latency multimodal assistance in real-world deployments, with a primary focus on blind and low-vision users.
OpenGlass resolves this gap with a split design: a lightweight glasses-side sensing unit captures visual context, while a nearby consumer-grade device performs multimodal inference and streams concise spoken feedback over a local Wi-Fi or hotspot link, avoiding reliance on remote cloud services.
We build on rapidly evolving multimodal models in the ecosystem, including large proprietary and open systems such as GPT-4~\cite{achiam2023gpt}, Gemini~\cite{team2023gemini}, and Qwen~\cite{bai2023qwen}, and we show how careful systems engineering enables practical assistive deployment on consumer hardware.
Meanwhile, there is also fast progress in compact on-device multimodal models, exemplified by MiniCPM-V and MiniCPM-o~\cite{minicpmv,yu2025minicpm}, which makes local deployment increasingly feasible at smaller footprints.

We introduce a system-level evaluation suite covering response quality, end-to-end latency from query completion to audio playback, and safety-first abstention with auditable logs.
OpenGlass supports representative interaction patterns for user-initiated visual assistance: obstacle and hazard awareness with concise action suggestions, user-initiated interruption of speech output with rapid task switching at the output layer, and scene-grounded follow-up queries in the current visual context.
These interaction patterns are evaluated under realistic wireless conditions and are designed to remain robust when the visual evidence is insufficient, in which case the system conservatively abstains and provides retake guidance.

\paragraph{System contributions.}
Three system-level designs are central to usability, privacy, and safety-awareness goals:
\begin{itemize}[leftmargin=*,nosep]
  \item \textbf{Split local deployment for responsive assistance.}
  We co-design a wearable sensing loop with nearby consumer-grade inference, enabling strong multimodal capability and responsive interaction under realistic local Wi-Fi or hotspot conditions while reducing default exposure of raw egocentric images to remote cloud services.
  \item \textbf{Streaming speech output with early feedback.}
  The system streams partial model outputs into speech synthesis as early as possible to reduce latency from query completion to the first audible response for the user.
  \item \textbf{System-level evaluation with auditable logs.}
  We provide an evaluation suite that measures response quality, latency, and safety-aware abstention, and we log timestamped interaction events and failure cases for reproducibility, privacy-conscious deployment analysis, and safety-oriented failure diagnosis.
\end{itemize}

\section{Approach}
\label{sec:approach}

\subsection{System Overview}

\begin{figure}[t]
\centering
\includegraphics[width=\linewidth]{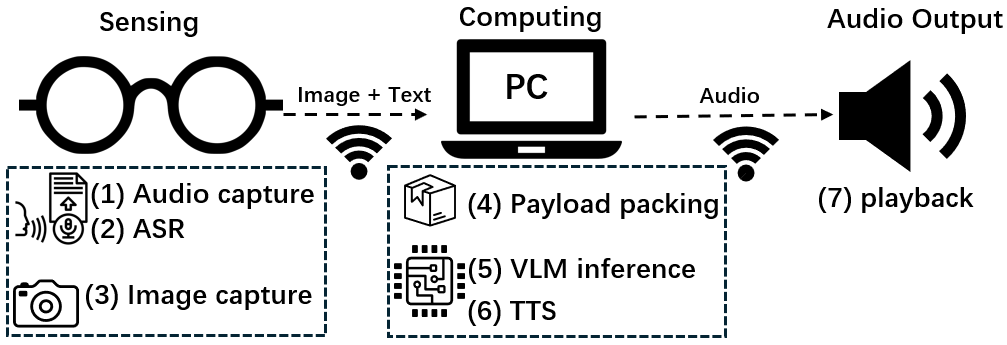}
\caption{System framework overview and latency decomposition. A glasses-side sensing unit captures audio and images, while a nearby consumer-grade device performs host-side packing, VLM inference, and TTS playback. Stages (1)--(7) define the end-to-end timeline used in our evaluation. Stage (7) performs speech synthesis on the host and outputs audio locally in our prototype; streaming the audio to a wearable speaker is an optional deployment choice.}
\label{fig:pipeline}
\end{figure}

\label{sec:approach_overview}
Concretely, OpenGlass operates as a streaming multimodal pipeline with early speech output.
After the user finishes a spoken query, the system first transcribes it with Automatic Speech Recognition (ASR), then captures a single frame from the wearable camera over Wi-Fi or cellular links.
On the host, we optionally preprocess the image (e.g., resizing and JPEG handling) and pack it together with the query into a request for the VLM backend (\texttt{llama.cpp}).
The VLM response is decoded in a streaming manner and segmented into speakable units (sentence-level flush), which are forwarded to TTS as early as possible to minimize the time to first audio playback.
Figure~\ref{fig:pipeline} summarizes the end-to-end stages and latency decomposition.

\paragraph{Wearable capture.}

The glasses-side camera runs ESP32-S3 \texttt{CameraWebServer}.
The host pulls an on-demand JPEG snapshot (\texttt{/capture}) for inference and optionally uses an MJPEG preview stream (\texttt{:81/stream}) for alignment.
The preview frame rate adapts to Wi-Fi throughput.

\paragraph{Edge inference and speech I/O.}

OpenGlass runs the VLM backend locally on an edge device (e.g., a Windows laptop with a consumer GPU).
In our current deployment, MiniCPM-V/o 4.5 is used in vision--text mode via \texttt{llama.cpp} runtime/server (image-conditioned response generation).
Since the GGUF + \texttt{llama.cpp} stack does not natively support omni audio tokens at the time of writing, we implement speech I/O modularly: Whisper for ASR and SAPI5 for TTS.
It preserves an end-to-end, demo-ready interaction loop while keeping the system extensible to future omni backends.

\subsubsection{Practical challenges}
\label{sec:approach_challenges}
Wearable assistive deployment faces three coupled issues: (1) low-cost sensors often produce noisy, blurred, or low-light images that can amplify hallucinations; (2) Wi-Fi capture and transport introduce latency and tail jitter that can dominate responsiveness; (3) fixed optics must serve both near-field reading and mid-range navigation, motivating proactive quality diagnosis (H1) and user guidance for retaking frames.

\subsubsection{Hardware}
\label{sec:approach_hw}
OpenGlass is designed around low-cost, off-the-shelf components: ESP32-S3 camera modules OV5640 on the glasses, and a commodity edge device (laptop/tablet/phone in future versions) for VLM inference and audio playback.
We emphasize this split architecture because it matches a practical glasses-to-phone deployment pattern while enabling reproducible experimentation.

\subsubsection{Software}
\label{sec:approach_sw}
OpenGlass consists of (1) a capture client that manages frame acquisition, resizing, and request packaging; (2) an edge VLM service based on \texttt{llama.cpp}; (3) streaming post-processing that segments partial outputs into speakable units (sentence-level flush); (4) an asynchronous TTS worker; and (5) logging utilities that record timestamps and interaction events for auditability. These timestamps define TTFT/TTFA and user$\rightarrow$audio consistently under both replay and Wi-Fi end-to-end measurements (Table~\ref{tab:wifi-e2e}).

\subsection{Key Design Choices}
\label{sec:approach_design}

\subsubsection{Problem statement}
\label{sec:approach_problem}
Current edge inference stacks provide strong pointwise VLM capability but lack native support for fully streaming, end-to-end multimodal interaction under tight latency budgets on commodity hardware.
In particular, full-duplex omni audio requires specialized runtimes and significantly higher hardware requirements, and is not reliably supported in our deployment environment.
Therefore, a robust solution must (1) maximize perceived responsiveness via streaming and parallelism, (2) remain safe under uncertainty and low-quality inputs, and (3) be modular so that components can be upgraded independently over time.

\paragraph{Our contribution.}
To fill this gap, we design a modular streaming middleware that composes wearable capture, VLM reasoning, and low-latency speech output into a demo-ready loop on consumer hardware.
It also supports systematic evaluation through auditable \textbf{quality-latency-safety} metrics and timestamped logs for reproducibility and failure analysis.

\subsubsection{Recommended implementation}
\label{sec:approach_impl}
We adopt a pragmatic and reproducible implementation path: (1) ESP32 provides camera frames (and optionally audio) over Wi-Fi; (2) the host runs ASR (faster-whisper) and the VLM+TTS pipeline; (3) the VLM response is streamed and spoken as early as possible.

\subsection{Blind-Assist Evaluation Set}
\label{sec:approach_dataset}

\subsubsection{Dataset introduction}
OpenGlass uses a task-oriented evaluation set tailored to time-critical assistive interaction.
It is intended as a scenario-grounded evaluation set rather than a large-scale, generic benchmark, with data captured directly from the ESP32 wearable camera to reflect realistic deployment conditions.
Many existing academic datasets are collected with high-quality sensors under controlled settings, whereas assistive wearables often face motion blur, low light, indoor clutter, and occlusions.
It contains 120 instances, each comprising (1) a single JPEG frame captured from ESP32 on demand, (2) a user query (typed or ASR-transcribed), and (3) a task label and rubric-based scores.
Instances are grouped into four capability families: T1 obstacle/hazard warning (30), T2 object finding and localization (30), T3 sign/QR understanding (30; T3A sign/text reading 24, T3B QR content 6), and H1 proactive vision with safe abstention (30).
This design enables comparable analysis of latency, success, abstention, and high-confidence errors across assistive tasks.

\subsubsection{Data collection}
Data collection follows the real system loop to ensure alignment with the demo:
frames are acquired via an on-demand snapshot interface, while an optional live preview stream is used to time user actions and select evaluation moments.
We intentionally cover diverse lighting and motion conditions, as well as varying Wi-Fi throughput and camera configurations (HD/SVGA/VGA), including both ``clean'' and ``hard'' frames to stress-test H1 (quality diagnosis and safe abstention).
We do not require dense pixel-level ground truth; instead, we score outputs using an interaction-centric rubric and report quality\_mean alongside success/abstain.

\subsubsection{Prompt design}
Prompting follows three principles: task explicitness, safety-first abstention, and output formatting.
We include the task ID (T1/T2/T3/H1) in the prompt and enforce task-specific output requirements:
T1 prioritizes risk conclusion and actionable guidance; T2 requires relative direction and next action; T3 only outputs text/QR content when reliably readable; H1 diagnoses input quality and suggests how to retake the frame.
To prevent harmful hallucinations, we hard-code abstention rules (e.g., never fabricate sign or QR content; never assert path safety without evidence).
Finally, responses follow a speakable structure (one-sentence action, one-sentence evidence/location; if abstaining, one-sentence reason plus 1--2 retake suggestions) to reduce TTS latency and facilitate rubric-based evaluation.

\paragraph{Prompt selection and standardization.}
We release the final prompt templates and per-instance prompt records to support reproducibility.

\section{Experiments and Results}
\label{sec:experiments}

\subsection{Experiment Setup}
\label{sec:exp_setup}
We evaluate OpenGlass on a commodity laptop (Lenovo Legion R9000P; AMD Ryzen 9 8945HX; NVIDIA GeForce RTX 5060 Laptop GPU with 8GB VRAM; 32GB RAM) running \texttt{llama.cpp}/\texttt{llama-server} with INT4-quantized MiniCPM-V 4.5 and MiniCPM-o 4.5 weights.
Wearable images are captured by an ESP32-S3 camera OV5640 module running \texttt{CameraWebServer} and retrieved on demand via snapshot endpoint. 

\subsection{Baselines}
\label{sec:exp_baselines}
We compare against (1) \textbf{Gemini 2.5 Flash}, a representative overseas cloud VLM API accessed via VPN (tail latency dominated by network); (2) \textbf{Qwen-VL-Max}, a chinese cloud VLM API; and (3) \textbf{NaiveOnDevice}, an unoptimized on-device pipeline that disables streaming, resizing, and safety, and waits for the full response before speaking.
Following Alibaba Cloud DashScope naming, \textbf{Qwen-VL-Max} corresponds to the strongest model in the Qwen2-VL series.

\subsection{Metrics}
\label{sec:exp_metrics}
We report latency, response quality, and safety-related behavior with consistent definitions. For backend responsiveness (Tables~\ref{tab:main_replay} and~\ref{tab:ablation_replay}), we measure \textbf{TTFT} (time from VLM request dispatch to the first generated token) and \textbf{TTFA$_\text{backend}$} (time from VLM request dispatch to the first audible output), evaluated under the disk-replay protocol and therefore excluding wearable capture and wireless transport.
For real deployment behavior (Table~\ref{tab:wifi-e2e}), we report \textbf{user$\rightarrow$audio} latency defined as the time from user query ready (ASR completed or text submitted) to the first audible output, which includes capture, packing, inference, and speech synthesis.

We score each response with a safety-first rubric \textbf{Quality}$\in\{0,1,2\}$: 2 = correct key content with actionable guidance; 1 = \emph{safe or partially correct behavior}, including conservative abstention with retake instructions when visual evidence is insufficient; 0 = incorrect or unsafe output, with fabricated sign/QR content or unsupported path-safety claims treated as high-risk errors.
We report \textbf{Success} as the fraction of outputs judged usable by the task-specific rubric, including conservative safe abstention when the task does not provide enough visual evidence. We report \textbf{Abstain} separately as the fraction of responses that explicitly choose a safe fallback, such as saying that the target is not confidently visible or asking the user to retake or scan the scene. Thus, Success and Abstain are not mutually exclusive: an abstention can be considered usable when it avoids a high-risk hallucination. We additionally report high-confidence error (HCE), which marks confident but unsafe or fabricated assertions.
For latency metrics, we report $p50$ (median) and $p95$ (95th percentile).

\subsection{Latency Definitions: Replay vs.\ Real Wi-Fi End-to-End}
\label{sec:exp_latency_defs}
To isolate algorithmic factors and enable reproducible ablations, \textbf{Tables~\ref{tab:main_replay}--\ref{tab:ablation_replay}} report \emph{replay latency} where the input image is loaded locally (disk replay) and thus excludes ESP32 capture and Wi-Fi transport.
To validate practical usability of the glasses pipeline, \textbf{Table~\ref{tab:wifi-e2e}} reports \emph{real end-to-end} latency that includes on-device capture and host-side request packing.
Thus, Table~\ref{tab:wifi-e2e} reflects true Wi-Fi end-to-end behavior, while other tables report replay latency for controlled comparison. We use replay for reproducibility and variable isolation (large-scale statistics and ablations), and additionally report Wi-Fi E2E to validate real usability.

In deployment, a practical median first-audio time can be approximated as TTFA (backend replay) plus Wi-Fi capture and host-side packing, noting that exact values depend on network and resolution. For example, combining medians from Table~\ref{tab:main_replay} (backend TTFA = 692\,ms) and Table~\ref{tab:wifi-e2e} (Wi-Fi capture = 401\,ms), and adding host-side packing overhead measured from our auditable logs ($\approx$15\,ms; i.e., resize/JPEG/base64/JSON), yields an estimated $\approx$1.11\,s from query-ready to first audio in deployment.

\begin{table}[t]
\centering
\tiny
\setlength{\tabcolsep}{2pt}
\begin{tabular}{lrrrrr}
\toprule
Method & TTFT$\downarrow$ & TTFA$\downarrow$ & Quality\,$\uparrow$ & Success\,$\uparrow$ & Abstain\,$\uparrow$ \\
\midrule
\multicolumn{6}{l}{\textbf{Cloud-based (remote inference)}} \\
\addlinespace[1.0pt]
\cdashline{1-6}[0.6pt/1.5pt]
\addlinespace[2.0pt]
Gemini 2.5 Flash & 5428 & 5441 & 1.425 & 0.958 & 0.317 \\
Qwen-VL-Max      & 616  & 984  & 1.408 & 0.933 & 0.300 \\
\midrule
\multicolumn{6}{l}{\textbf{Local on-device (local inference)}} \\
\addlinespace[1.0pt]
\cdashline{1-6}[0.6pt/1.5pt]
\addlinespace[2.0pt]
NaiveOnDevice          & 1380 & 1381 & 1.333 & 0.900 & 0.325 \\
MiniCPM-o 4.5 (Ours)   & \textbf{518} & \textbf{682} & 1.283 & \textbf{0.908} & 0.317 \\
MiniCPM-o 4.5 Think (Ours) & 544 & 710 & 1.017 & 0.850 & 0.217 \\
MiniCPM-V 4.5 (Ours)   & 526 & 692 & \textbf{1.325} & 0.892 & \textbf{0.375} \\
\bottomrule
\end{tabular}
\captionsetup{font=small}
\caption{Main results under the disk-replay protocol (local images; $n=120$), isolating backend latency from camera capture and wireless transport. \textbf{TTFT/TTFA} are in ms. \textbf{Quality} is a safety-first rubric score in \{0,1,2\}: 2 indicates a correct actionable response, 1 indicates safe or partially correct behavior such as abstention with retake guidance, and 0 indicates incorrect or unsafe output (e.g., fabricated sign/QR content). \textbf{Success} is the task-specific usable-output rate and may include safe abstention when evidence is insufficient. \textbf{Abstain} is the explicit safe-fallback rate.}
\label{tab:main_replay}
\end{table}

\subsection{Main Results (Replay Latency)}
\label{sec:exp_main}
Table~\ref{tab:main_replay} compares major baselines under the replay protocol (n=120).
OpenGlass achieves sub-second backend responsiveness under the replay protocol (TTFT/TTFA in the 0.5--0.7s range), while the overseas cloud baseline exhibits multi-second delays.
The domestic cloud baseline is faster than overseas, yet remains slower than our edge pipeline on TTFA.
Across MiniCPM-V and MiniCPM-o backends, latency is comparable, indicating that enabling an omni-trained model as a vision--text backend does not introduce a noticeable latency penalty in this deployment regime.

\paragraph{Task-level quality breakdown.}
Table~\ref{tab:task-breakdown-ours-v} shows that quality is task-dependent rather than uniformly low.
H1 self-checking is strongest (Q=2.000, HCE=0), and T1 hazard awareness has no HCE but often falls back conservatively.
The main bottlenecks are T3A/T3B: sign reading has 37.5\% HCE, and the small QR subset has the lowest quality.
T2 mostly abstains, avoiding hallucinated target localization.
These results motivate abstention, tool-based QR decoding, sign verification, and multi-frame confirmation.

\begin{table}[t]
\centering
\small
\begin{tabular}{lrrrrr}
\toprule
Task & $n$ & Quality & Success & Abstain & HCE \\
\midrule
T1 & 30 & 1.233 & 100.0\% & 46.7\% & 0.0\% \\
T2 & 30 & 1.000 & 100.0\% & 100.0\% & 0.0\% \\
T3A & 24 & 1.250 & 62.5\% & 0.0\% & 37.5\% \\
T3B & 6 & 0.333 & 33.3\% & 16.7\% & 66.7\% \\
H1 & 30 & 2.000 & 100.0\% & 0.0\% & 0.0\% \\
\midrule
\textbf{All} & 120 & 1.325 & 89.2\% & 37.5\% & 10.8\% \\
\bottomrule
\end{tabular}
\caption{Per-task breakdown of \textbf{Ours\,(V)} on the v6 benchmark ($n{=}120$). Quality is the rubric mean (0--2). Success is the task-specific usable-output rate and may include safe abstention. Abstain is the explicit safe-fallback rate. HCE denotes high-confidence errors.}
\label{tab:task-breakdown-ours-v}
\end{table}

\subsection{Ablation Study (Replay Latency)}
\label{sec:exp_ablation}
Table~\ref{tab:ablation_replay} shows that \textbf{resizing is the dominant latency factor}:
disabling resize roughly doubles TTFT/TTFA (526/692$\rightarrow$1220/1379\,ms), while resizing to 448px yields a large speedup (178/335\,ms) with a modest quality drop.
Streaming mainly improves perceived responsiveness; disabling streaming increases latency and may also affect measured TTFT due to different buffering/dispatch behavior.

\begin{table}[t]
\centering
\footnotesize
\begin{tabular}{lrrr}
\toprule
Ablation & TTFT$_{P50}\,\downarrow$ & TTFA$_{P50}\,\downarrow$ & Quality$\uparrow$ \\
\midrule
Full & 526 & 692 & 1.33 \\
w/o Streaming & 715 & 731 & 1.32 \\
w/o ParallelTTS & 521 & 725 & 1.32 \\
w/o Resize & 1220 & 1379 & 1.33 \\
w/o Safety & 506 & 676 & 1.33 \\
Resize448 & 178 & 335 & 1.29 \\
\bottomrule
\end{tabular}
\captionsetup{font=small}
\caption{Ablations under replay latency (local images).}
\label{tab:ablation_replay}
\end{table}

\subsection{Real Wi-Fi End-to-End Latency (Glasses Pipeline)}
\label{sec:exp_wifi_e2e}
Table~\ref{tab:wifi-e2e} reports end-to-end latency over the real ESP32 Wi-Fi camera link.
Our full pipeline (capture + resize + on-device VLM + local TTS) achieves a median user-to-audio latency of \textbf{993\,ms} with a 97.5\% below-2 s pass rate.
Resizing the 1280$\times$720 capture to 896$\times$504 cuts TTFT from 1222 to 528\,ms (2.3$\times$ speedup).
Cloud alternatives are far slower: Qwen-VL-Max reaches 2065\,ms median (46.7\% pass rate) and Gemini~2.5 Flash exceeds 4.9\,s (0\% pass rate), with P95 tails above 7\,s and 11\,s respectively.
Results support median sub-2 s user-initiated assistance, while also showing that long-tail Wi-Fi and TTFT jitter must be handled with retake guidance and explicit abstention.

\begin{table}[t]
\centering
\tiny
\setlength{\tabcolsep}{3pt}
\begin{tabular}{lcccc}
\toprule
\textbf{Metric} & \textbf{Ours} & \textbf{Ours (Raw)} & \textbf{Qwen-VL-Max} & \textbf{Gemini 2.5 Flash} \\
\midrule
VLM Input Size       & 896$\times$504 & 1280$\times$720 & 896$\times$504 & 896$\times$504 \\
Capture P50 (ms)     & 401   & 371   & 385   & 414   \\
TTFT P50 (ms)        & \textbf{528}  & 1222  & 1675  & 4483  \\
TTFT P95 (ms)        & \textbf{1167} & 1596  & 6778  & 11086 \\
user$\rightarrow$audio P50 (ms) & \textbf{993}  & 1625  & 2065  & 4927  \\
user$\rightarrow$audio P95 (ms) & \textbf{1780} & 2041  & 7293  & 11475 \\
$<$2s Pass Rate      & \textbf{97.5\%} & 93.3\% & 46.7\% & 0.0\% \\
\bottomrule
\end{tabular}
\captionsetup{font=small}
\caption{Wi-Fi end-to-end latency comparison. All systems use ESP32 Wi-Fi camera capture with pyttsx3 local TTS. Latency is in milliseconds.}
\label{tab:wifi-e2e}
\end{table}

\paragraph{Failure attribution.}
We further attribute the below-2 s failures in Table~\ref{tab:wifi-e2e}.
Local TTS is never the bottleneck: the failure counts (Wi-Fi / inference-or-API / TTS) are 3 (2/2/0) for resized OpenGlass, 8 (0/8/0) for raw OpenGlass, 64 (6/64/0) for Qwen-VL-Max, and 120 (--/120/0) for Gemini 2.5 Flash.
Counts are not mutually exclusive because a slow trial can combine capture and inference tails.
Thus, the rare failures of resized OpenGlass come from ESP32 capture jitter and TTFT tails, raw-payload failures come from VLM inference tails induced by larger 1280$\times$720 inputs, and cloud failures are dominated by API/inference latency.
This motivates timeout handling, retake guidance, adaptive resolution control, and conservative abstention for long-tail cases.

\subsection{Resolution--Throughput Trade-off}
\label{sec:exp_resolution}
Table~\ref{tab:resolution_tradeoff} shows the resolution--latency trade-off for dynamic camera control.
Lower resolutions reduce payload size and MiniCPM-V slices: VGA reaches 537 ms mean TTFT with 3 slices, while HD uses 7 slices and reaches 1150 ms.
Thus we downshift from HD to SVGA/VGA under weak Wi-Fi.

\begin{table}[t]
\centering
\scriptsize
\setlength{\tabcolsep}{2.5pt}
\renewcommand{\arraystretch}{0.92}
\resizebox{\columnwidth}{!}{%
\begin{tabular}{lrrrrrrr}
\toprule
Res. & Size & Wi-Fi(ms) & JPEG(KB) & Payload(KB) & TTFT(ms) & Slices &
\begin{tabular}[c]{@{}c@{}}User$\rightarrow$Audio\\(mean/P95, ms)\end{tabular} \\
\midrule
HD   & 1280$\times$720 & 342 & 58.6 & 78.1 & 1150 & 7 & 1493 / 1596 \\
XGA  & 1024$\times$768 & 400 & 57.6 & 76.8 & 1176 & 5 & 1577 / 1678 \\
SVGA & 800$\times$600  & 281 & 33.9 & 45.2 & 893 & 5 & 1175 / 1286 \\
VGA  & 640$\times$480  & 283 & 24.4 & 32.5 & 537 & 3 & 825 / 903 \\
\bottomrule
\end{tabular}%
}
\captionsetup{font=small}
\caption{Resolution sweep (averages). ``Slices'' is the number of MiniCPM-V image slices used by the backend.}
\label{tab:resolution_tradeoff}
\end{table}

\subsection{Key Takeaways}
\label{sec:exp_takeaways}
\textbf{(1) Local deployment latency advantage.} The on-device pipeline provides substantially lower and more predictable responsiveness than cloud baselines, especially under overseas access.
\textbf{(2) Resize dominates.} Lightweight resizing is the main lever for reducing both backend and end-to-end latency.
\textbf{(3) Wearable responsiveness is practical.} Including ESP32 capture and Wi-Fi transport, the pipeline achieves high below-2 s pass rates for query-ready, user-initiated visual assistance. The failure attribution above shows that local TTS is not the bottleneck; the remaining long-tail cases are dominated by capture jitter and inference/API tails. This motivates explicit abstention, retake guidance, adaptive resolution control, and timeout handling in safety-sensitive deployments.

\subsection{Interactive Robustness}
\label{sec:exp_interactive}
Beyond raw latency, assistive glasses require robust interaction patterns such as interruption barge-in.
We report system-level barge-in tests in the supplementary material (Table~\ref{tab:bargein}): OpenGlass achieves millisecond-scale stop latency with 100\% success and zero carry-over.
We also evaluate MiniCPM-o "think" strategies (Tables~\ref{tab:think_analysis}) and find that enabling thinking is detrimental under short-response constraints; thus we recommend disabling thinking for real-time assistive deployment.

\paragraph{Rubric-based evaluation and auditable logs.}
We evaluate outputs with a safety-first 0/1/2 rubric: a score of 2 requires correct key facts with actionable guidance; a score of 1 credits safe or partially correct behavior, including abstention accompanied by concrete retake instructions; and a score of 0 penalizes high-risk errors, especially hallucinated sign/QR content or unsafe navigation claims.
For transparency, we maintain auditable interaction logs at the instance level, including the task label, the prompt template used, the reference answer, and the model’s raw response.
These records allow reviewers and future users to trace failures to specific prompts and conditions, reproduce qualitative examples, and analyze trade-offs between quality, latency, abstention, and high-confidence errors.

\paragraph{Use cases and demo setup.}
We demonstrate OpenGlass as a complete, deployable edge pipeline for blind and low-vision assistance, emphasizing privacy-oriented on-device inference and real-time usability across three interaction patterns: walk-and-talk obstacle awareness, scene glance with uncertainty-aware clarification, and interrupt-and-switch barge-in. The wearable camera runs ESP32-S3 \texttt{CameraWebServer}; the host uses MJPEG preview for alignment and pulls on-demand JPEG frames for inference, with local VLM inference and modular ASR/TTS. See Appendix~\ref{sec:demo_screenshots} for the wearable UI and live terminal instrumentation.

\paragraph{Availability and reproducibility.}
OpenGlass is released at \url{https://github.com/OpenSQZ/OpenGlass} under an open-source license.
The release includes ESP32 firmware/configuration, PC client scripts, \texttt{llama.cpp} server commands, MiniCPM-V / MiniCPM-o vision-text configs, prompts, manifests, raw logs, aggregation scripts, and the CSV files for Tables~\ref{tab:main_replay}--\ref{tab:wifi-e2e}.
The default setup uses an ESP32-S3 OV5640 camera, an RTX 5060 laptop, local \texttt{llama.cpp}, 1280$\times$720 capture, 896$\times$504 VLM input, and local \texttt{pyttsx3} TTS.

\section{Related Work}
\label{sec:related}

\subsection{VLM and OLM}
Assistive visual question answering and human-in-the-loop systems such as VizWiz~\cite{bigham2010vizwiz} demonstrate the value of visual assistance for blind and low-vision users, while recent cloud multimodal assistants and agentic systems further improve visual understanding and natural-language interaction~\cite{huang2025my}.
These systems motivate our application setting, but cloud-centered or human-in-the-loop assistance is not designed around local-first privacy, real wireless end-to-end latency, and automatic safety-aware abstention.

Vision-language models (VLMs) enable open-ended image understanding and instruction following, exemplified by LLaVA-style visual instruction tuning~\cite{liu2023visual,liu2024improved} and compact open-source families such as MiniCPM-V~\cite{minicpmv}.
Omni-modal language models (OLMs) further unify audio, vision, and text, aiming for streaming spoken interaction and, in some cases, full-duplex dialogue~\cite{defossez2024moshi,xie2024mini,minicpmo45}.
These models highlight the promise of ``see-and-say'' assistants, but deploying them for safety-sensitive assistive use requires system-level handling of uncertainty, interruption, latency tails, and auditable behavior beyond model accuracy alone.

\subsection{VLM/OLM for Local Compute Platform}
Recent work explores making VLMs practical on resource-constrained hardware via efficient architectures and deployment stacks, such as MobileVLM~\cite{wu2024mobilevlm} and surveys on small/edge VLMs~\cite{sharshar2025vision,patnaik2025small}.
In parallel, quantized inference runtimes like \texttt{llama.cpp} make local deployment on commodity devices increasingly accessible~\cite{llamacpp}.
However, model-centric edge deployment typically reports backend inference behavior, while a wearable assistive system must also account for camera capture, Wi-Fi transport, request packing, streaming speech output, and safety behavior under uncertain visual evidence.

OpenGlass complements prior assistive, cloud, and edge-VLM systems by combining local-first egocentric privacy, wearable ESP32 sensing, real Wi-Fi E2E latency, safety-aware quality/HCE evaluation, and open-source artifacts.
Our contribution is therefore not the split architecture alone, but a practical and auditable end-to-end co-design for local MLLM-driven visual assistance.

\section{Conclusion}
In this work, we introduce \textsc{OpenGlass}, an open-source, privacy-oriented, local-first system for low-latency multimodal visual assistance, with a primary focus on blind and low-vision users.
OpenGlass follows a sensing-computing split design: a lightweight glasses-side unit captures visual context, while a nearby consumer-grade device performs local VLM inference and local speech output, reducing default reliance on remote cloud services and limiting exposure of raw egocentric visual data.

Our experiments show that such a local-first pipeline can provide practical responsiveness under real ESP32 Wi-Fi capture.
With resized visual payloads, OpenGlass achieves 993 ms median user-to-audio latency and a 97.5\% below-2 s pass rate; with raw 1280$\times$720 payloads, it achieves 1625 ms median latency and a 93.3\% below-2 s pass rate.
The results also show that lightweight resizing and the resulting reduction in VLM image slices are major latency drivers, while safety-aware prompts and auditable logs help identify abstentions and high-risk hallucinations.

OpenGlass should be interpreted as a user-initiated visual assistance reference platform rather than a certified navigation aid.
It can provide obstacle/hazard awareness, sign and object assistance, image-quality self-checks, and interruptible spoken feedback, but it is not intended to replace canes, guide dogs, or professional mobility tools in safety-critical navigation.
Future work includes BLV user studies, multi-frame confirmation for dynamic hazards, adaptive resolution control under weak wireless conditions, and stronger privacy controls for egocentric recording and logging.

\section*{Limitations}
\label{sec:limitations}

\paragraph{Scope of assistance.}
OpenGlass is a research prototype and reference implementation for user-initiated visual assistance, not a certified navigation or mobility aid.
Although the system provides obstacle/hazard awareness and concise action suggestions, it should not be used as a replacement for canes, guide dogs, human orientation-and-mobility training, or other professional assistive tools.
The current prototype is best suited to query-driven tasks such as scene glances, object finding, sign understanding, image-quality self-checking, and interruptible spoken feedback.
It is not designed to guarantee safety in fast dynamic situations such as street crossing, dense crowds, moving vehicles, or rapidly changing obstacles.

\paragraph{Latency tails and wireless reliability.}
Our Wi-Fi end-to-end results show median sub-2 s response times, but the metric is a pass rate rather than a hard real-time guarantee.
Even in the best reported setting, 97.5\% of trials fall below 2 s, which means that a small fraction of interactions can still exceed the target due to ESP32 capture jitter, Wi-Fi congestion, TTFT tails, or local runtime scheduling.
For safety-sensitive use, such long-tail cases should trigger conservative behavior: abstain when the evidence is insufficient, ask the user to slow down or retake the frame, and avoid issuing confident motion guidance when the scene is unclear.
Future versions should include stronger adaptive resolution control, local hotspot/Wi-Fi Direct support, timeouts and retries, and multi-frame confirmation before presenting dynamic hazards as actionable.

\paragraph{Vision and hardware limits.}
The current low-cost OV5640 camera and fixed optics impose a practical sensing boundary.
Our evaluation emphasizes large signs, nearby objects, coarse obstacle/hazard awareness, and image-quality diagnosis; fine-grained small-text OCR, distant signage, low-light motion, and dynamic outdoor navigation remain challenging.
The sensing-computing split also uses a nearby consumer-grade host in our experiments, so the paper should be read as a reproducible edge-host reference platform rather than a glasses-only embedded deployment.
Lower-power mobile SoCs, NPUs, and fully integrated audio output remain important future deployment targets.

\paragraph{Evaluation and user-study limits.}
Our 120-instance benchmark is scenario-grounded rather than a full accessibility user study.
Task-conditioned prompts intentionally isolate hazard awareness, object finding, sign/QR understanding, and image-quality self-checking, making latency--quality--safety trade-offs auditable, but may overestimate open-ended deployment where the system must infer user intent.
We leave unified intent routing, BLV participant studies, expert feedback, inter-annotator agreement, and longitudinal field testing to future work.

\section*{Ethical Considerations}
\label{sec:ethics}
\paragraph{Privacy and egocentric data.}
First-person cameras may capture bystanders, screens, documents, private homes, workplaces, or other sensitive scenes.
OpenGlass reduces exposure by defaulting to local VLM inference and local speech output, but this is a privacy-oriented deployment choice rather than a formal privacy guarantee.
Deployments should minimize retention of raw frames and spoken content, disable or anonymize raw-image logging outside evaluation, avoid releasing identifiable bystander imagery, and provide users with clear controls over recording and deletion.
If cloud APIs or online TTS are enabled as optional enhancements, the interface should clearly indicate that data may leave user-controlled devices.

\section*{Acknowledgments}
This work was funded by the Shanghai Qi Zhi Institute Innovation Program (SQZ202410).

\bibliography{custom}

\begin{thebibliography}{16}
\providecommand{\natexlab}[1]{#1}

\bibitem[{Achiam et~al.(2023)Achiam, Adler, Agarwal, Ahmad, Akkaya, Aleman,
  Almeida, Altenschmidt, Altman, Anadkat et~al.}]{achiam2023gpt}
Josh Achiam, Steven Adler, Sandhini Agarwal, Lama Ahmad, Ilge Akkaya,
  Florencia~Leoni Aleman, Diogo Almeida, Janko Altenschmidt, Sam Altman,
  Shyamal Anadkat, and 1 others. 2023.
\newblock Gpt-4 technical report.
\newblock \emph{arXiv preprint arXiv:2303.08774}.

\bibitem[{Bai et~al.(2023)Bai, Bai, Chu, Cui, Dang, Deng, Fan, Ge, Han, Huang
  et~al.}]{bai2023qwen}
Jinze Bai, Shuai Bai, Yunfei Chu, Zeyu Cui, Kai Dang, Xiaodong Deng, Yang Fan,
  Wenbin Ge, Yu~Han, Fei Huang, and 1 others. 2023.
\newblock Qwen technical report.
\newblock \emph{arXiv preprint arXiv:2309.16609}.

\bibitem[{Bigham et~al.(2010)Bigham, Jayant, Ji, Little, Miller, Miller,
  Miller, Tatarowicz, White, White et~al.}]{bigham2010vizwiz}
Jeffrey~P Bigham, Chandrika Jayant, Hanjie Ji, Greg Little, Andrew Miller,
  Robert~C Miller, Robin Miller, Aubrey Tatarowicz, Brandyn White, Samual
  White, and 1 others. 2010.
\newblock Vizwiz: nearly real-time answers to visual questions.
\newblock In \emph{Proceedings of the 23nd annual ACM symposium on User
  interface software and technology}, pages 333--342.

\bibitem[{D{\'e}fossez et~al.(2024)D{\'e}fossez, Mazar{\'e}, Orsini, Royer,
  P{\'e}rez, J{\'e}gou, Grave, and Zeghidour}]{defossez2024moshi}
Alexandre D{\'e}fossez, Laurent Mazar{\'e}, Manu Orsini, Am{\'e}lie Royer,
  Patrick P{\'e}rez, Herv{\'e} J{\'e}gou, Edouard Grave, and Neil Zeghidour.
  2024.
\newblock Moshi: a speech-text foundation model for real-time dialogue.
\newblock \emph{arXiv preprint arXiv:2410.00037}.

\bibitem[{ggml org()}]{llamacpp}
ggml org.
\newblock llama.cpp: Llm inference in c/c++.
\newblock \url{https://github.com/ggml-org/llama.cpp}.
\newblock GitHub repository, accessed 2026-02-26.

\bibitem[{Huang et~al.(2025)Huang, Zhang, Liu, Qin, Zhu, Naumann, Chen, and
  Poon}]{huang2025my}
James~Y Huang, Sheng Zhang, Qianchu Liu, Guanghui Qin, Tinghui Zhu, Tristan
  Naumann, Muhao Chen, and Hoifung Poon. 2025.
\newblock Be my eyes: Extending large language models to new modalities through
  multi-agent collaboration.
\newblock \emph{arXiv preprint arXiv:2511.19417}.

\bibitem[{Liu et~al.(2024)Liu, Li, Li, and Lee}]{liu2024improved}
Haotian Liu, Chunyuan Li, Yuheng Li, and Yong~Jae Lee. 2024.
\newblock Improved baselines with visual instruction tuning.
\newblock In \emph{Proceedings of the IEEE/CVF conference on computer vision
  and pattern recognition}, pages 26296--26306.

\bibitem[{Liu et~al.(2023)Liu, Li, Wu, and Lee}]{liu2023visual}
Haotian Liu, Chunyuan Li, Qingyang Wu, and Yong~Jae Lee. 2023.
\newblock Visual instruction tuning.
\newblock \emph{Advances in neural information processing systems},
  36:34892--34916.

\bibitem[{OpenBMB()}]{minicpmo45}
OpenBMB.
\newblock Minicpm-o 4.5: An open-source omni multimodal model.
\newblock \url{https://github.com/OpenBMB/MiniCPM-o}.
\newblock GitHub repository (release v4.5), accessed 2026-02-26.

\bibitem[{Patnaik et~al.(2025)Patnaik, Nayak, Agrawal, Khamaru, Bal, Panda,
  Raj, Meena, and Vadlamani}]{patnaik2025small}
Nitesh Patnaik, Navdeep Nayak, Himani~Bansal Agrawal, Moinak~Chinmoy Khamaru,
  Gourav Bal, Saishree~Smaranika Panda, Rishi Raj, Vishal Meena, and Kartheek
  Vadlamani. 2025.
\newblock Small vision-language models: A survey on compact architectures and
  techniques.
\newblock \emph{arXiv preprint arXiv:2503.10665}.

\bibitem[{Sharshar et~al.(2025)Sharshar, Khan, Ullah, and
  Guizani}]{sharshar2025vision}
Ahmed Sharshar, Latif~U Khan, Waseem Ullah, and Mohsen Guizani. 2025.
\newblock Vision-language models for edge networks: A comprehensive survey.
\newblock \emph{IEEE Internet of Things Journal}.

\bibitem[{Team et~al.(2023)Team, Anil, Borgeaud, Alayrac, Yu, Soricut,
  Schalkwyk, Dai, Hauth, Millican et~al.}]{team2023gemini}
Gemini Team, Rohan Anil, Sebastian Borgeaud, Jean-Baptiste Alayrac, Jiahui Yu,
  Radu Soricut, Johan Schalkwyk, Andrew~M Dai, Anja Hauth, Katie Millican, and
  1 others. 2023.
\newblock Gemini: a family of highly capable multimodal models.
\newblock \emph{arXiv preprint arXiv:2312.11805}.

\bibitem[{Wu et~al.(2024)Wu, Xu, Liu, Tan, Liujianfeng, Li, Luan, Wang, and
  Shang}]{wu2024mobilevlm}
Qinzhuo Wu, Weikai Xu, Wei Liu, Tao Tan, Liujian Liujianfeng, Ang Li, Jian
  Luan, Bin Wang, and Shuo Shang. 2024.
\newblock Mobilevlm: A vision-language model for better intra-and inter-ui
  understanding.
\newblock In \emph{Findings of the Association for Computational Linguistics:
  EMNLP 2024}, pages 10231--10251.

\bibitem[{Xie and Wu(2024)}]{xie2024mini}
Zhifei Xie and Changqiao Wu. 2024.
\newblock Mini-omni2: Towards open-source gpt-4o with vision, speech and duplex
  capabilities.
\newblock \emph{arXiv preprint arXiv:2410.11190}.

\bibitem[{Yao et~al.(2024)Yao, Yu, Zhang, Wang, Cui, Zhu, Cai, Li, Zhao, He
  et~al.}]{minicpmv}
Yuan Yao, Tianyu Yu, Ao~Zhang, Chongyi Wang, Junbo Cui, Hongji Zhu, Tianchi
  Cai, Haoyu Li, Weilin Zhao, Zhihui He, and 1 others. 2024.
\newblock Minicpm-v: A gpt-4v level mllm on your phone.
\newblock \emph{arXiv preprint arXiv:2408.01800}.

\bibitem[{Yu et~al.(2025)Yu, Wang, Wang, Huang, Ma, He, Cai, Chen, Huang, Zhao
  et~al.}]{yu2025minicpm}
Tianyu Yu, Zefan Wang, Chongyi Wang, Fuwei Huang, Wenshuo Ma, Zhihui He,
  Tianchi Cai, Weize Chen, Yuxiang Huang, Yuanqian Zhao, and 1 others. 2025.
\newblock Minicpm-v 4.5: Cooking efficient mllms via architecture, data, and
  training recipe.
\newblock \emph{arXiv preprint arXiv:2509.18154}.

\end{thebibliography}
\clearpage
\appendix

\section{Interactive Robustness and Think Strategy (Supplementary)}
\label{sec:supp_interactive}

\subsection{System-level Barge-in (Interrupt) Test}
\label{sec:supp_bargein}
We evaluate barge-in at the \emph{pipeline level}: on user interrupt, the system should stop speech, clear pending TTS buffers, and resume a higher-priority query without leaking content from the interrupted utterance.
Table~\ref{tab:bargein} reports results across three trigger delays.
OpenGlass achieves 100\% success with near-instant stop (median $T_{\text{stop}}\approx 3.5$\,ms), low switching overhead ($T_{\text{ovhd}}\approx 98$\,ms), and \emph{zero carry-over} errors.

\begin{table}[!t]
\centering
\scriptsize
\setlength{\tabcolsep}{3pt}
\renewcommand{\arraystretch}{0.90}
\resizebox{\columnwidth}{!}{%
\begin{tabular}{lrrrrrr}
\toprule
Delay & $n$ & Succ. &
$T_{\text{stop}}$ &
$T_{\text{ovhd}}$ &
$T_{\text{resume}}$ &
Carry-over \\
 &  &  & P50 & P50 & P50 &  \\
\midrule
500\,ms  & 20 & 100\% & 3.6 & 102 & 1\,039 & 0\% \\
1\,000\,ms & 20 & 100\% & 3.5 &  96 &   893 & 0\% \\
2\,000\,ms & 20 & 100\% & 3.4 &  97 &   973 & 0\% \\
\midrule
All      & 60 & 100\% & 3.5 &  98 &   976 & 0\% \\
\bottomrule
\end{tabular}}
\captionsetup{font=small}
\caption{Barge-in interruption test.
All times in ms (P50).
$T_{\text{stop}}$: interrupt to speech halt.
$T_{\text{ovhd}}$: system switching cost (TTS teardown + re-init).
$T_{\text{resume}}$: interrupt to new audio, dominated by VLM inference for the replacement query.
Carry-over: leaked content from the interrupted utterance.}
\label{tab:bargein}
\end{table}

\subsection{Think Strategy is Detrimental under Short-response Constraints}
\label{sec:supp_think}
MiniCPM-o provides an optional ``think'' mode, but in short-response assistive settings it is consistently harmful.
Table~\ref{tab:think_analysis} shows Think ON reduces rubric quality (1.283$\rightarrow$1.017) and success rate (90.8\%$\rightarrow$85.0\%), and an adaptive rule does not recover quality.
The largest regression occurs on sign/text reading (T3A), where thinking increases hallucinations and high-confidence errors.

\begin{table}[!t]
\centering
\scriptsize
\setlength{\tabcolsep}{4pt}
\renewcommand{\arraystretch}{0.95}
\begin{tabular}{lrrr}
\toprule
\multicolumn{4}{l}{\textbf{Panel A: Overall Think Strategy}} \\
\midrule
Strategy & Quality(\textbf{Q}) $\uparrow$ & Success $\uparrow$& Abstain $\uparrow$ \\
\midrule
Think OFF & 1.283 & 90.8\% & 31.7\% \\
Think ON  & 1.017 & 85.0\% & 21.7\% \\
Adaptive  & 1.125 & 85.0\% & 33.3\% \\
\midrule
\multicolumn{4}{l}{\textbf{Panel B: Per-task Impact of Think}} \\
\midrule
Task & Think OFF \textbf{Q} & Think ON \textbf{Q} & $\Delta$\textbf{Q} \\
\midrule
H1 (quality) & 1.867 & 1.500 & -0.37 \\
T1 (hazard)  & 1.200 & 1.133 & -0.07 \\
T2 (find)    & 1.000 & 1.000 & +0.00 \\
T3A (sign)   & 1.083 & 0.333 & -0.75 \\
T3B (QR)     & 1.000 & 0.833 & -0.17 \\
\bottomrule
\end{tabular}
\captionsetup{font=small}
\caption{Think strategy analysis (Supplementary). Panel A compares overall strategies; Panel B reports per-task impact.}
\label{tab:think_analysis}
\end{table}

\begin{figure}[t]
\centering
\includegraphics[width=\linewidth]{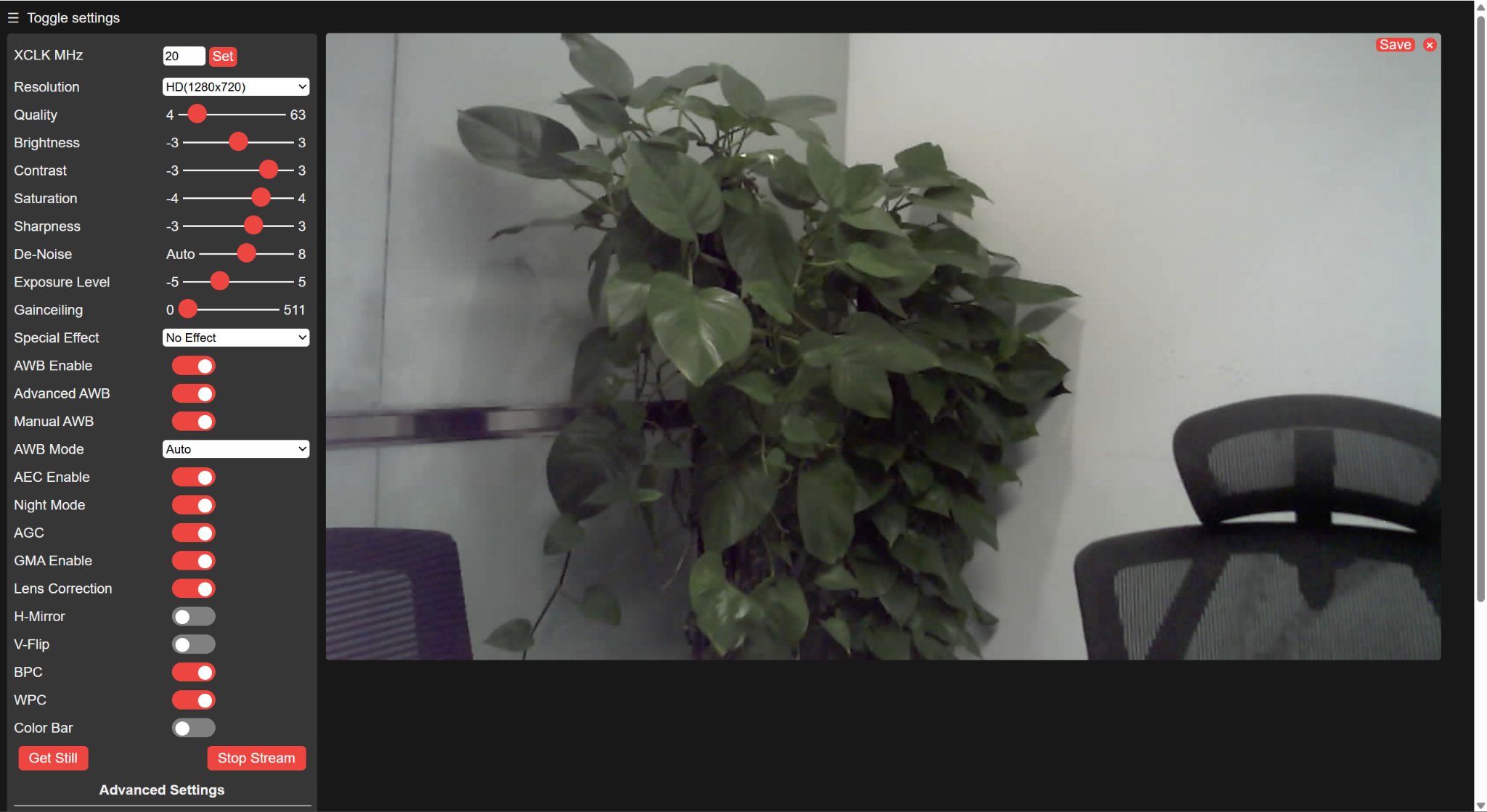}
\captionsetup{font=small}
\caption{CameraWebServer control panel and MJPEG preview on the ESP32-S3/OV5640 wearable camera. OpenGlass pulls an on-demand JPEG via on-demand snapshot interface for inference while optionally using live preview stream for preview and alignment.}
\label{fig:camerawebserver}
\end{figure}

\begin{figure}[t]
\centering
\includegraphics[width=\linewidth]{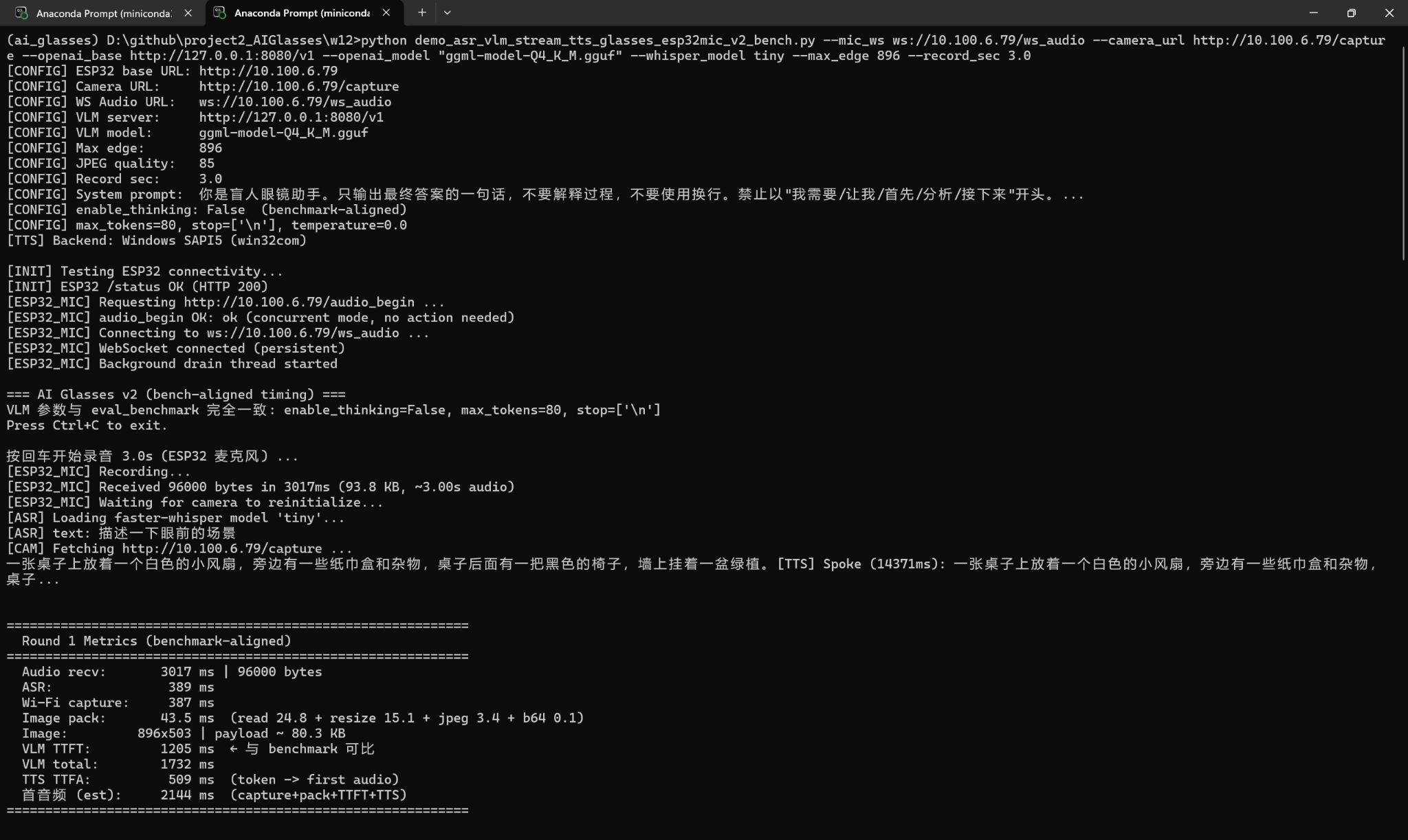}
\captionsetup{font=small}
\caption{Live OpenGlass run in an Anaconda prompt showing end-to-end instrumentation (audio receive, ASR, Wi-Fi capture, image packing, VLM TTFT/total, and TTS). These auditable logs support both reproducible evaluation and failure analysis.}
\label{fig:terminal_demo}
\end{figure}

\section{Demo Screenshots}
\label{sec:demo_screenshots}
Figure~\ref{fig:camerawebserver} shows the wearable camera UI, and Figure~\ref{fig:terminal_demo} shows the live end-to-end instrumentation.
\begin{table*}[t]
\centering
\small
\setlength{\tabcolsep}{5pt}
\renewcommand{\arraystretch}{1.10}
\begin{tabularx}{\textwidth}{l X X X}
\toprule
\textbf{Task} & \textbf{Prompt goal} & \textbf{Output format (spoken)} & \textbf{Safety / abstention rules} \\
\midrule
T1 & One sentence: Hazard/obstacle warning within 3m (left/center/right). &
One sentence: hazard + action (e.g., stop/slow/turn). &
If uncertain: safe abstain + retake guidance; never claim "safe to go" when unsure. \\

T2 & One sentence: Find a target object and provide relative direction. &
One sentence: found/not-found + direction + next action. &
If not confident: say not sure + suggest scanning; avoid false positive "found". \\

T3A & One sentence: Read the most salient sign text and explain its meaning. &
One sentence: "Sign: \texttt{<text>}, meaning: ..." or "unreadable". &
\textbf{No fabrication:} hallucinated text $\rightarrow 0$; unreadable $\rightarrow$ abstain with retake guidance. \\

T3B & One sentence: Detect a QR code and decode its payload. &
One sentence: "QR: \texttt{<payload>}" or "cannot decode". &
\textbf{No fabrication:} hallucinated payload/link $\rightarrow 0$; cannot decode $\rightarrow$ abstain with retake guidance. \\

H1 & One sentence: Diagnose image quality (ok/dim/blur) and give advice. &
One sentence: quality + actionable fix. &
Encourage proactive guidance (light/closer/steady); do not overclaim readability. \\
\bottomrule
\end{tabularx}
\caption{Task-conditioned prompting summary. We use a safety-first rubric where safe abstention can score 1 and hallucinated sign/QR content is penalized.}
\label{tab:prompt_summary}
\end{table*}

\begin{figure*}[t]
\centering
\includegraphics[width=0.24\textwidth]{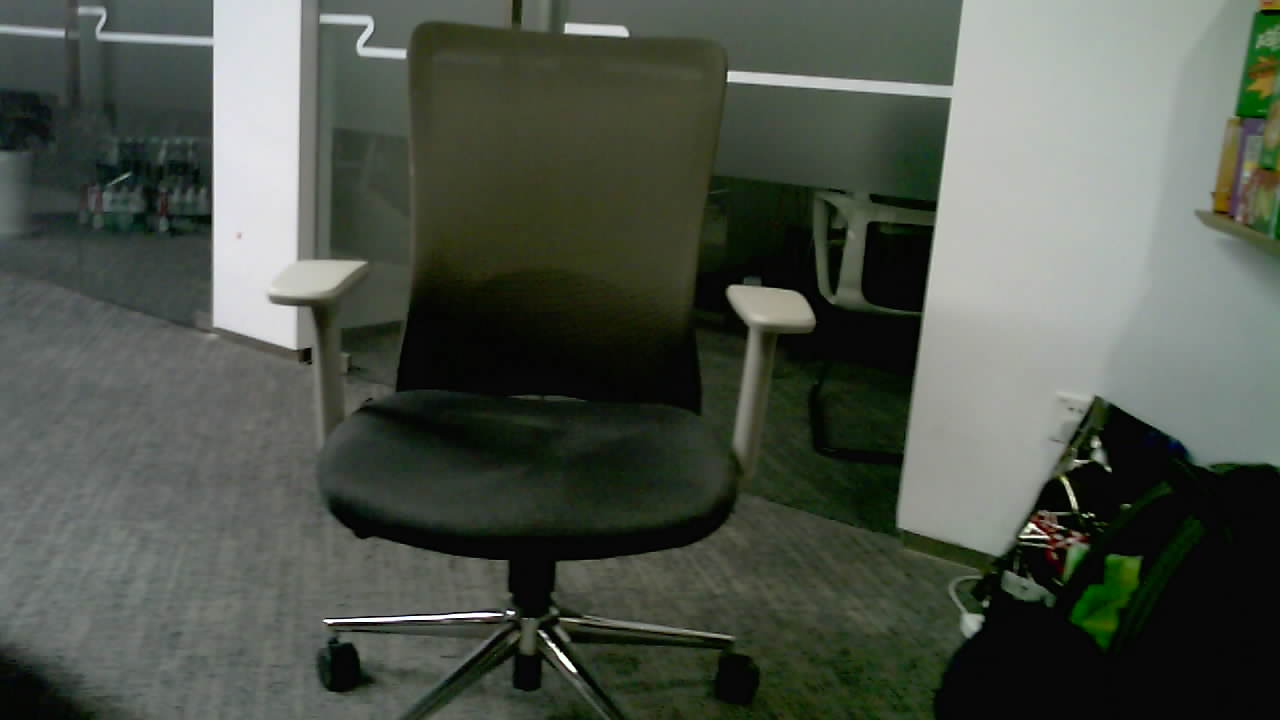}
\includegraphics[width=0.24\textwidth]{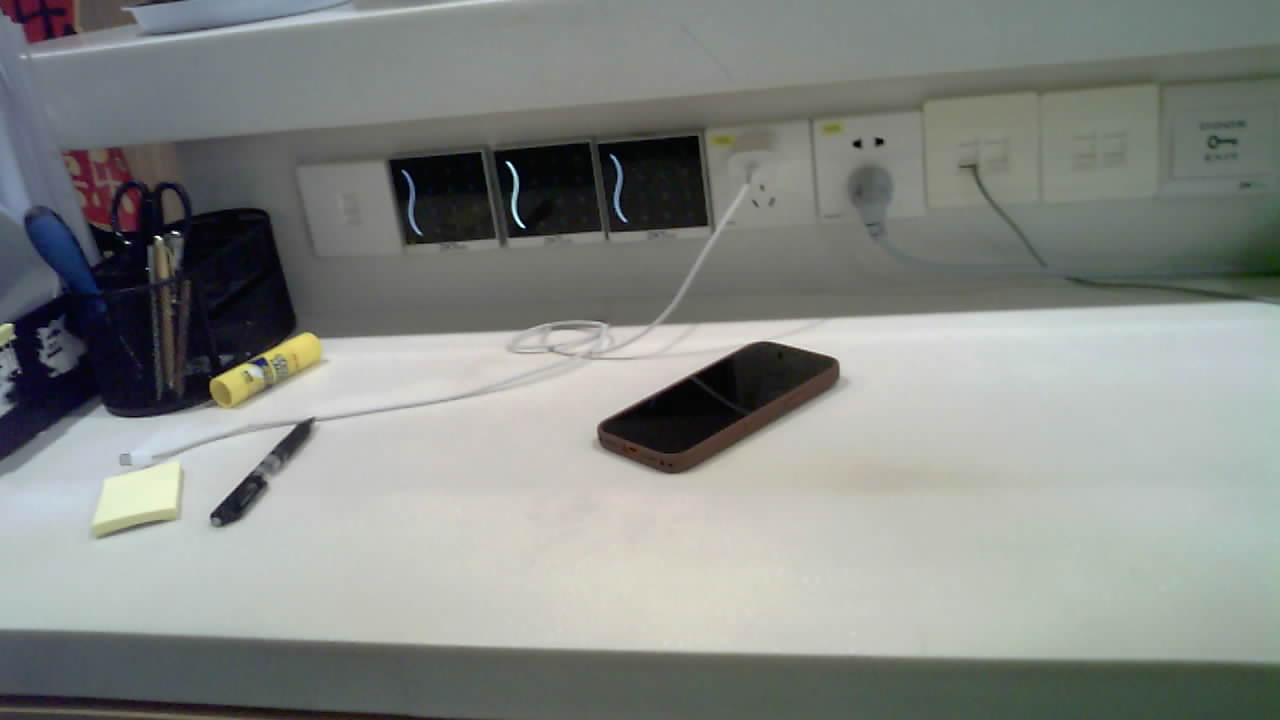}
\includegraphics[width=0.24\textwidth]{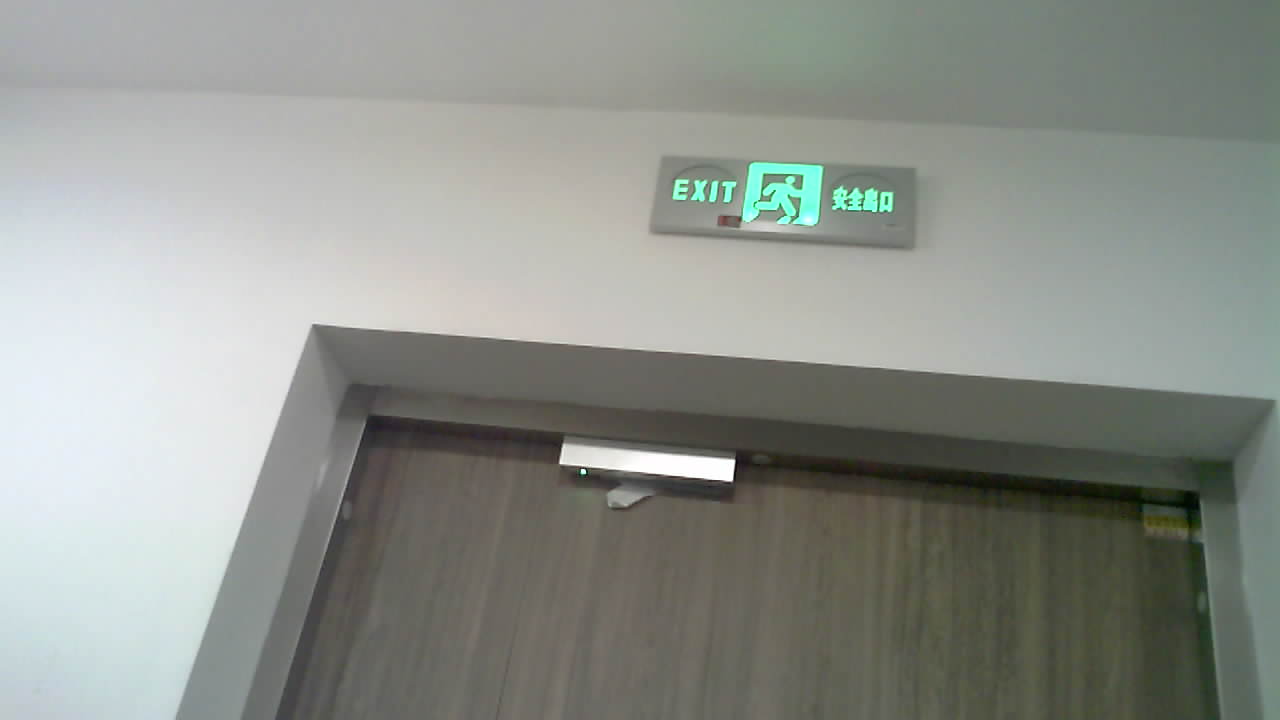}
\includegraphics[width=0.24\textwidth]{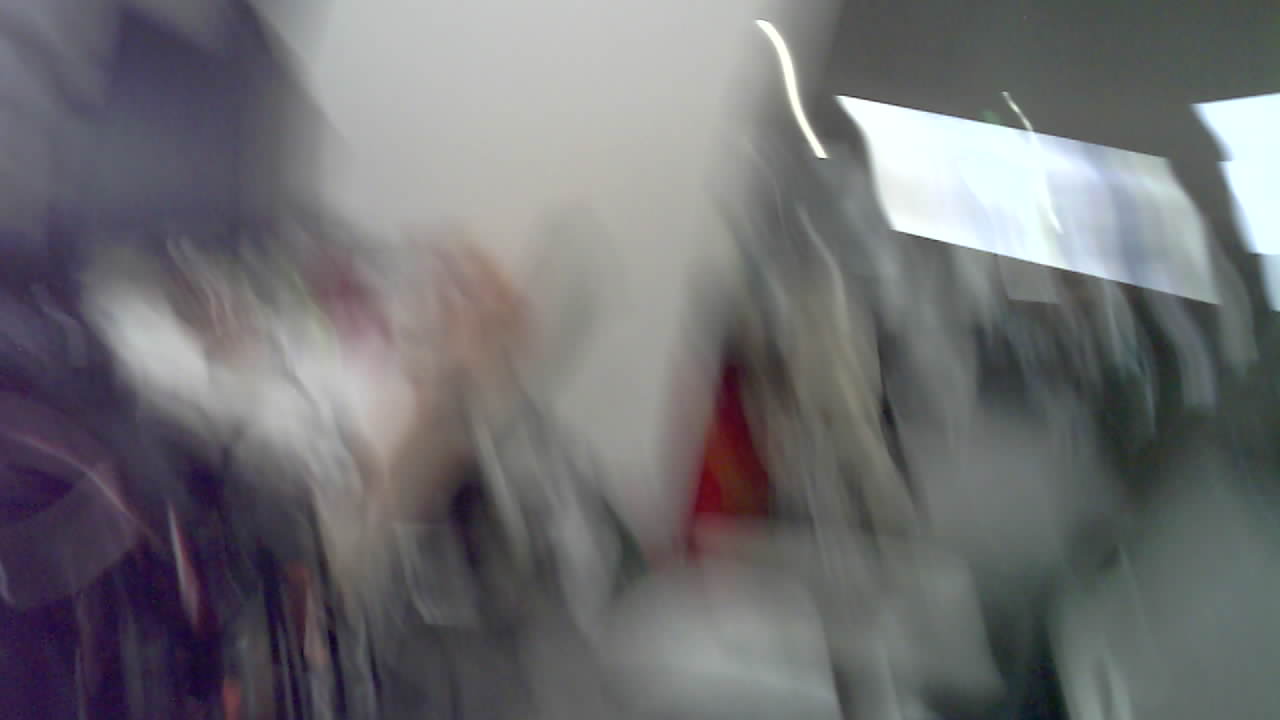}
\caption{Representative evaluation frames for each subtask (cropped from 1280$\times$720 ESP32 captures): (T1) obstacle warning, (T2) object finding, (T3A) sign reading, and (H1) quality diagnosis/safe abstention.}
\label{fig:task_examples}
\end{figure*}

\noindent\textbf{Representative inputs.}
Figure~\ref{fig:task_examples} shows representative ESP32 frames for our subtasks, illustrating the visual conditions that the prompts in Table~\ref{tab:prompt_summary} are designed to handle.
We include both "clean" task frames (T1/T2/T3A) and a low-quality example (H1) to motivate safety-first abstention and retake guidance when the input is unreliable.

\begin{figure*}[b]
\centering
\includegraphics[width=0.32\textwidth]{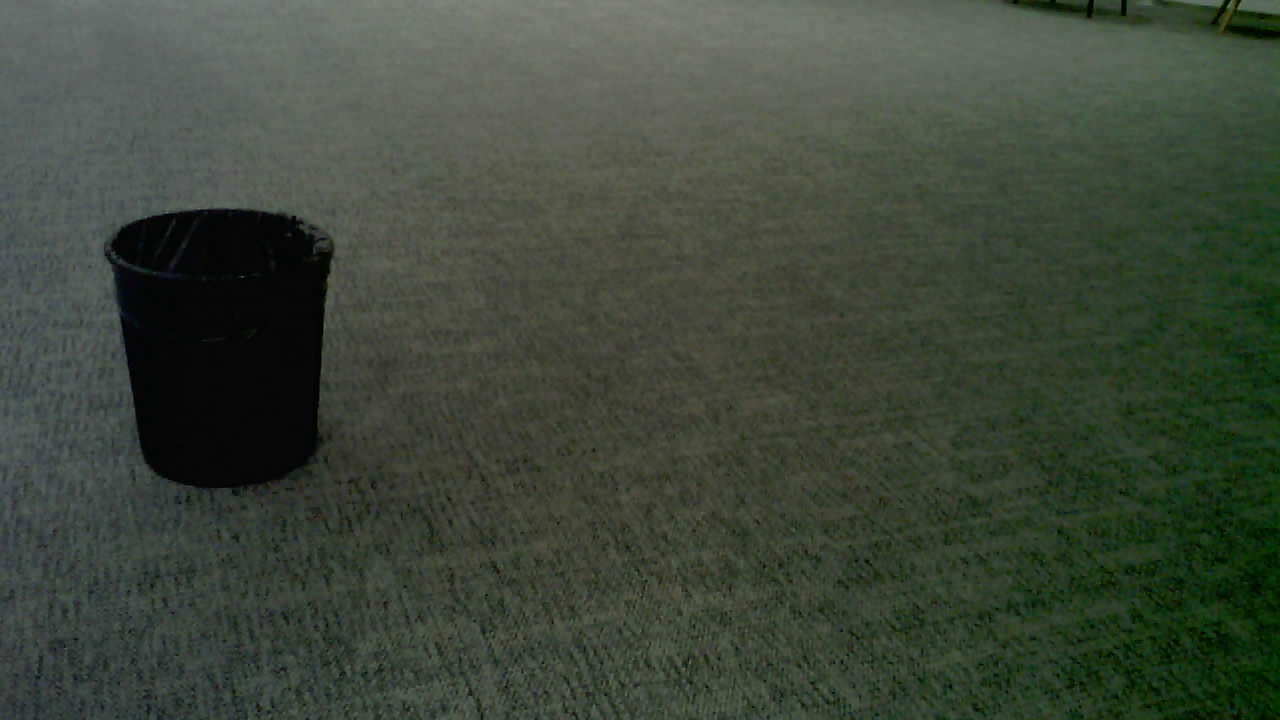}
\includegraphics[width=0.32\textwidth]{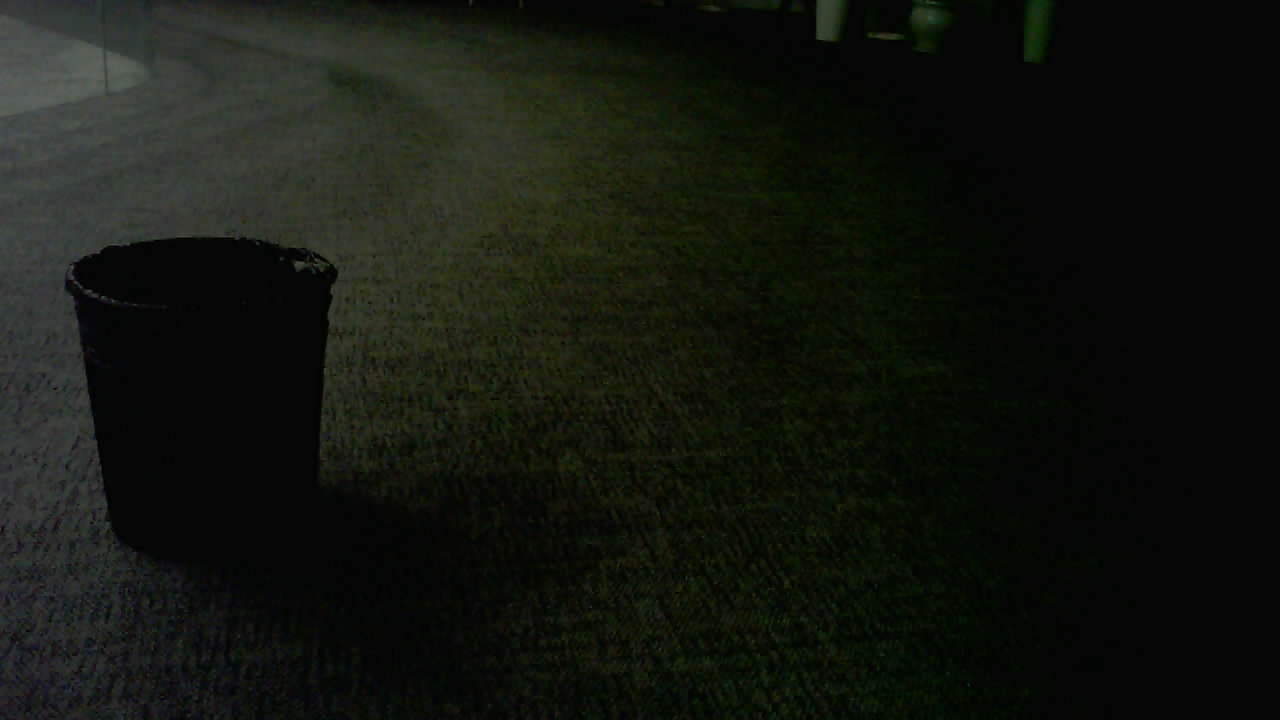}
\includegraphics[width=0.32\textwidth]{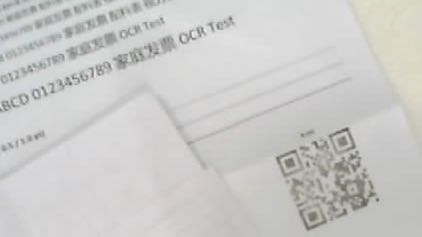}
\caption{Qualitative inputs corresponding to the examples in the text: (left) T1 walk-and-talk obstacle scene, (middle) T1 low-quality frame triggering safe abstention, (right) T3B QR frame where hallucinated payloads can occur. Images are cropped from 1280$\times$720 ESP32 captures for visibility.}
\label{fig:qual_examples}
\end{figure*}

\paragraph{Qualitative examples. (Fig.~\ref{fig:qual_examples})}
\begin{itemize}[leftmargin=*,nosep]
\item \textbf{Walk-and-talk (T1).} "There is a trash can ahead; it is recommended to go around to the right." (fast actionable guidance).
\item \textbf{Safe abstention (T1).} "If you can't see clearly or are unsure, please hold still and reshoot." (safety-first fallback under uncertainty).
\item \textbf{Failure mode (T3B).} For QR decoding, the model may hallucinate a URL; our rubric penalizes fabricated payloads and motivates strict abstention rules.
\end{itemize}

\end{document}